\documentclass[letterpaper, 10 pt, conference]{ieeeconf} 
\IEEEoverridecommandlockouts
\usepackage{mathrsfs}
\usepackage{xcolor}

\usepackage{graphicx} 
\usepackage{graphics} 
\usepackage{epsfig} 
\usepackage{amsmath} 
\usepackage{amssymb} 
\usepackage{color}
\usepackage{bm}
\usepackage{multirow}
\usepackage{array}
\usepackage{booktabs}
\usepackage{algorithm,algcompatible}
\usepackage[noend]{algpseudocode}
\usepackage{mathtools}
\usepackage{cite}
\usepackage{makecell}
\usepackage{hyperref}
\usepackage{cuted}
\usepackage{balance}
\definecolor{pointcolor}{RGB}{0,114,189}
\definecolor{linecolor}{RGB}{217,83,25}
\definecolor{planecolor}{RGB}{237,177,32}
\definecolor{spherecolor}{RGB}{119,172,48}
\definecolor{ellipsoidcolor}{RGB}{126,47,142}
\definecolor{cylindercolor}{RGB}{77,190,238}
\definecolor{conecolor}{RGB}{162,20,47}

\graphicspath{{figures/}}

\title{\huge \bf LEAR: Learning Edge-Aware Representations for Event-to-LiDAR Localization}

\author{Kuangyi Chen\quad  Jun Zhang\quad Yuxi Hu\quad Yi Zhou\quad Friedrich Fraundorfer
\thanks{Kuangyi Chen, Jun Zhang, Yuxi Hu, and Friedrich Fraundorfer are with the Institute of Visual Computing, Graz University of Technology, 8010 Graz, Austria. E-mail: \tt\small\{kuangyi.chen, jun.zhang, yuxi.hu,  friedrich.fraundorfer\}@tugraz.at}
\thanks{Yi Zhou is with the Neuromorphic Automation and Intelligence Lab
(NAIL), School of Artificial Intelligence and Robotics, Hunan University, Changsha 410012, China. E-mail: \tt\small\ eeyzhou@hnu.edu.cn}}

\begin{document}
\maketitle
\begin{abstract}
Event cameras offer high-temporal-resolution sensing that remains reliable under high-speed motion and challenging lighting, making them promising for localization from LiDAR point clouds in GPS-denied and visually degraded environments. However, aligning sparse, asynchronous events with dense LiDAR maps is fundamentally ill-posed, as direct correspondence estimation suffers from modality gaps. We propose LEAR, a dual-task learning framework that jointly estimates edge structures and dense event–depth flow fields to bridge the sensing-modality divide. Instead of treating edges as a post-hoc aid, LEAR couples them with flow estimation through a cross-modal fusion mechanism that injects modality-invariant geometric cues into the motion representation, and an iterative refinement strategy that enforces mutual consistency between the two tasks over multiple update steps. This synergy produces edge-aware, depth-aligned flow fields that enable more robust and accurate pose recovery via Perspective-n-Point (PnP) solvers. On several popular and challenging datasets, LEAR achieves superior performance over the best prior method. The source code, trained models, and demo videos are made publicly available online\footnote{https://github.com/EasonChen99/LEAR}.

\end{abstract}

\section{Introduction}
Visual localization aims to estimate the six degrees of freedom (6-DoF) pose of a camera within a reference map. It is a promising and widely adopted approach, as cameras are lightweight and cost-effective, making them particularly well-suited for deployment on compact robotic platforms such as Unmanned Aerial Vehicles (UAVs). However, standard cameras are susceptible to motion blur and lighting variations. Event cameras, a new class of bio-inspired sensors, offer compelling advantages—including high dynamic range, low latency, and asynchronous operation—which make them robust to such challenges. Consequently, event cameras have attracted growing interest as an alternative to standard cameras for visual localization. 

A core challenge in event-based localization lies in establishing correspondences between events and reference maps, of which the signal or data representations are significantly different, with events usually being sparse but texture-aware, while point clouds typically being dense and geometrically informative. Traditional approaches often rely on task-specific reference maps, such as geo-tagged event frames\cite{fischer2020event, 9635907, nair2024enhancing}, handcrafted feature databases \cite{milford2015towards, fischer2022many}, or learned descriptors \cite{kong2022event, hussaini2024applications, claxton2024improving}. While being effective for place recognition, these approaches typically yield coarse localization and often require large effort to build and maintain. To achieve more accurate localization, other methods construct 3D line maps \cite{yuan2016fast} or semi-dense reconstructions \cite{yuan2024evit, zuo2024cross, xu2025mets}, aligning events with projected map edges. However, their performance suffers under noise, occlusion, or weak texture. Photometric map-based methods \cite{gallego2017event, bryner2019event} attempt to match predicted intensity changes to observed events, but they lack view-dependent appearance modeling, struggling in dynamic or complex environments. Direct pose regression methods \cite{nguyen2019real, jin20216, lin20226, ren2024simple} offer efficient alternatives by implicitly encoding scene geometry within network parameters. However, they exhibit limited generalization to novel environments.

\begin{figure}[t]
    \centering
    \includegraphics[width=0.95\linewidth]{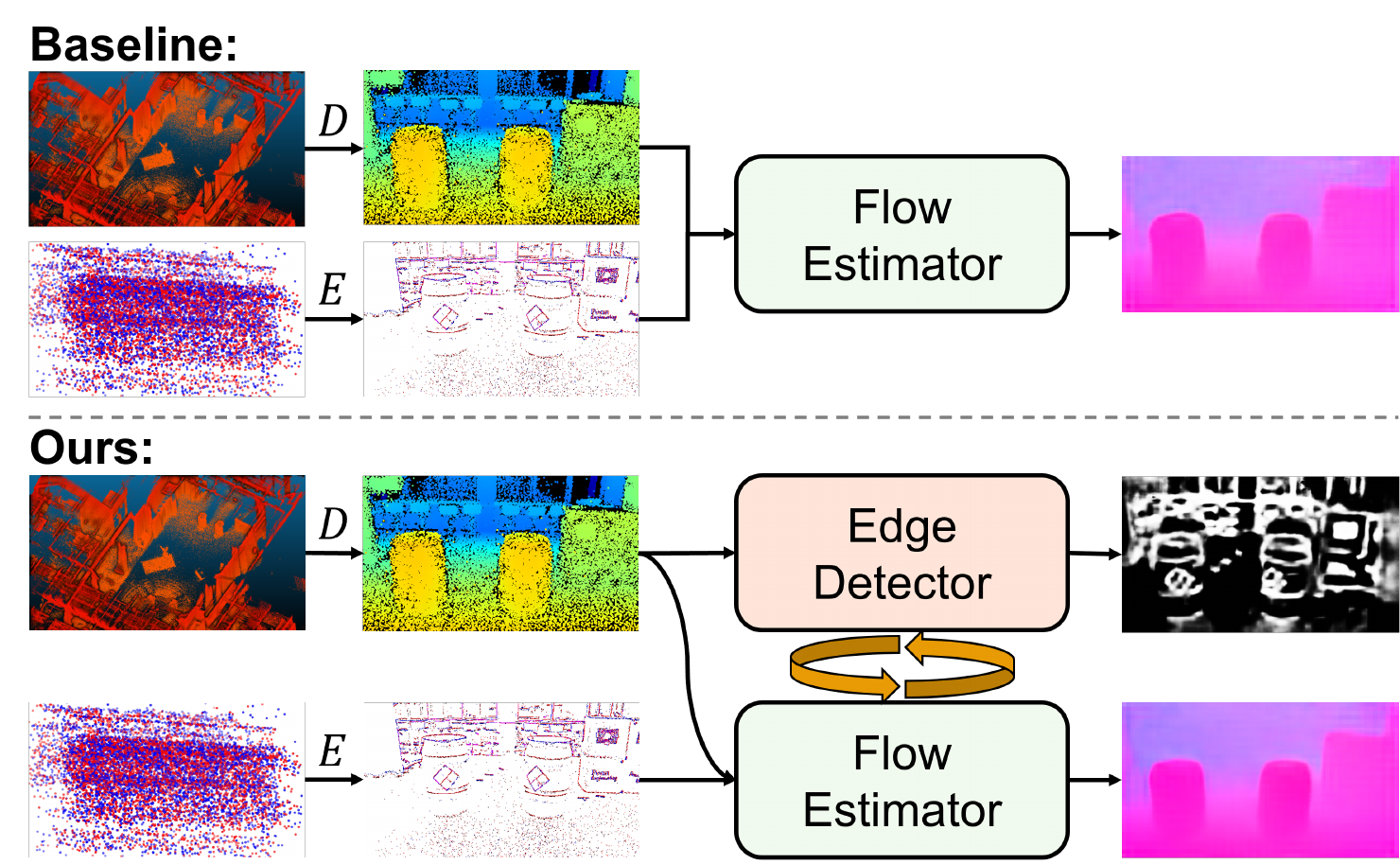}
    \caption{\emph{Our method} (\textbf{bottom}) integrates a flow estimator with an edge detector within a mutually reinforcing cycle, enabling more accurate 2D–3D matching (i.e., event–depth flow) compared to the flow-only baseline \emph{EVLoc}\cite{chen2025evloc} (\textbf{top}).} 
    \label{fig:Teaser}
    \vspace{-5mm}
\end{figure}

Unlike these task-specific representations, LiDAR point clouds offer a compelling alternative as reference maps: they are geometrically precise, robust to lighting, and often readily available via third-party providers. However, directly aligning events with LiDAR remains challenging due to their inherently different nature. EVLoc \cite{chen2025evloc}, to our knowledge, makes the first attempt to bridge this gap by formulating the problem as dense flow estimation between events and projected LiDAR points in 2D space. The flow estimates are then used to determine 2D-3D correspondences for camera pose recovery. While promising results are demonstrated, the approach functions largely as a black box, offering limited interpretability into the underlying mechanism.

Our key insight is that edge geometry forms a modality-invariant bridge between events and LiDAR: events are triggered at intensity gradients, while depth discontinuities in LiDAR capture structural boundaries. Incorporating this shared structure into the representation can reduce the burden on cross-modal feature encoding and make correspondence estimation more robust.
In our case, masking LiDAR point clouds using event-derived edge maps (as shown in Fig.~\ref{fig:Feature}, \emph{Ground Truth Edge Map}) leads to significantly improved performance compared to using raw point clouds alone, as demonstrated by the \emph{oracle} setting in Table~\ref{tab:Ablation}. However, generating such aligned edge masks requires knowledge of the relative pose—i.e., the associated correspondences we seek to estimate. This highlights a mutual dependency: accurate edge extraction depends on known correspondences, while good-quality correspondences require well-aligned, edge-aware encoded features.

Motivated by this insight, we introduce LEAR, a dual-task framework for event-based localization using LiDAR maps, as shown in Fig.~\ref{fig:Teaser}. 
Given an initial pose guess—obtained through coarse global localization methods such as visual place recognition—LEAR refines it to the precise 6-DoF camera pose.
LEAR couples a dense event–depth flow estimator with an edge detector in a mutually reinforcing architecture. A Cross-task Feature Fusion (CFF) module embeds modality-invariant edge cues into the motion estimation pipeline at the feature level, improving correspondence reliability across modalities. An Iterative Feature Refinement (IFR) module then recurrently updates both edge and flow features, enforcing geometric consistency and progressively improving prediction quality. This synergy yields edge-aware, depth-aligned flow fields that produce more stable and accurate pose recovery via a RANSAC-based PnP solver.
Overall, our main contributions are summarized as follows:
\begin{itemize}
    \item We propose a novel dual-task learning framework for event-based visual localization from LiDAR maps, which integrates a flow estimator and an edge detector, enabling edge-aware representations that enhance cross-modal consistency for event-LiDAR data alignment.
    \item We introduce the CFF and IFR modules to facilitate the learning of both task branches in a mutually reinforcing manner, improving localization and edge prediction.
    \item We conduct extensive experiments on public datasets of varying scenarios, and demonstrate the effectiveness of our proposed edge feature enhancement modules and state-of-the-art performance in localization accuracy.
\end{itemize}

\section{Related Work}
Event-based visual localization has been approached from multiple perspectives. In this section, we review key developments along the mainstream directions.

Retrieval methods avoid cross-modal matching by converting event streams into queryable representations, such as images\cite{fischer2020event, milford2015towards, nair2024enhancing} or edge maps\cite{lee2021eventvlad}, enabling similarity-based matching against a reference database. Others construct handcrafted\cite{fischer2022many} or learned descriptors\cite{kong2022event, claxton2024improving}, or encode implicit representations in network weights\cite{hussaini2024applications}. However, these approaches typically require costly database construction and maintenance, and achieve only coarse localization, which can serve as an initialization step for our method.

Another line of work focuses on training neural networks to regress camera poses directly from events\cite{nguyen2019real, jin20216, lin20226, ren2024simple}, encoding environmental information implicitly. For example, Nguyen et al.\cite{nguyen2019real} estimate poses from short event bursts using a CNN and Stacked Spatial LSTM. Although these methods eliminate explicit maps, they suffer from limited scalability and poor generalization to unseen environments.

Pose refinement approaches assume a known 3D scene representation and an initial estimate of the event camera’s state. Based on the type of scene representation, these approaches can be broadly categorized into geometric and photometric methods. Geometric methods typically construct a 3D map enriched with edge features and optimize the camera pose by aligning events with these features. For instance, Yuan et al. \cite{yuan2016fast} build a scene map consisting solely of vertical lines and achieve 3-DoF localization by matching 2D event lines with their 3D counterparts. More recent approaches \cite{yuan2024evit, zuo2024cross, xu2025mets} construct semi-dense maps using depth sensors or Structure from Motion (SfM) techniques, and estimate 6-DoF poses through polarity-aware geometric alignment between events and the semi-dense maps. However, since these methods rely heavily on explicit edge features, their performance degrades in scenes with noisy or weak texture. Photometric methods infer intensity changes from scene maps that contain color information \cite{gallego2017event, bryner2019event, liu2024gs}. For example, Bryner et al. \cite{bryner2019event} propose a method that estimates poses by aligning observed and predicted intensity change images, using a generative event model and a photometric 3D map. However, traditional photometric maps typically lack view-dependent appearance modeling, limiting their expressiveness under varying viewpoints. To address this, Liu et al. \cite{liu2024gs} introduce a Gaussian Splatting-based map, which is able to encode more complex ambient lighting effects. Nevertheless, rendering artifacts or inconsistencies in Gaussian Splatting can directly impact localization accuracy.

Compared to the aforementioned approaches, LiDAR maps offer dense, accurate, and lighting-invariant geometry, making them attractive for event-based localization. Additionally, LiDAR maps are often readily available from third-party providers, eliminating the need for task-specific scene construction. Despite this potential, few methods have addressed the challenge of cross-modal matching between events and LiDAR point clouds. \cite{ta2023l2e, xing2023target} are among the first attempts to handle this for event–LiDAR extrinsic calibration. However, they typically rely on task-specific markers or controlled environments, limiting their scalability.
E2PNet~\cite{lin2023e2pnet} introduces a learning-based, image-like event feature representation and applies it to existing RGB-based methods~\cite{li2021deepi2p, pham2020lcd} for event-to-point cloud localization. However, it reports only coarse localization performance.
The most closely related work with our method is EVLoc\cite{chen2025evloc}, which tackles alignment by directly learning dense flow fields between events and LiDAR maps. However, it doesn't explicitly exploit modality-invariant features that can strengthen cross-modal correspondence. 
In contrast, we propose a dual-task framework that explicitly targets this representational bottleneck by jointly learning an edge detector and a flow estimator in a mutually reinforcing manner, enabling consistency in feature space and significantly improving event–LiDAR localization robustness and accuracy.  
\section{Preliminaries}
\label{sec:method}
In this section, we describe the common modules inherited from the baseline EVLoc~\cite{chen2025evloc}, including data preprocessing and the flow estimation network.

\begin{figure*}[!ht]
    \centering
    \includegraphics[width=1.0\linewidth]{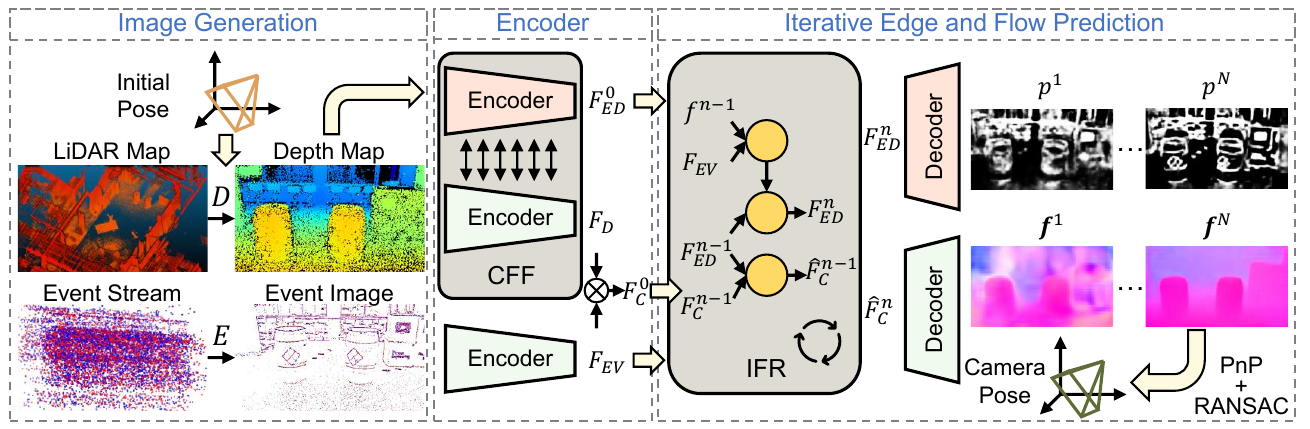}
    \caption{\textit{The three stages of the proposed method}: 1) \textit{Image Generation}: a depth map is generated using a pinhole camera model $D$ based on a given initial pose guess, and an event image is constructed from the raw event stream using an event image generator $E$; 2) \textit{Encoder}: the depth map and event image are fed into the flow estimation branch (\textbf{green module}), and the depth map also serves as input to the edge detection branch (\textbf{red module}). Encoded depth features $F_D$ and edge features $F_{ED}$ are fused across scales via the CFF module to produce edge-aware representations; 3) \textit{Iterative Edge and Flow Prediction}: these, along with event features $F_{EV}$ and correlation volumes $F_C$ (constructed by indexing the element-wise products of $F_D$ and $F_{EV}$ using the current flow; see Fig.~\ref{fig:IFR}), are passed to the IFR module, where flow $\boldsymbol{f}$ and edge representations $F_{ED}$ are refined iteratively and jointly, ultimately producing flow estimates $\boldsymbol{f}^{1\rightarrow N}$ and edge estimates $p^{1\rightarrow N}$ from the corresponding decoders. The final flow estimate $\boldsymbol{f}^{N}$ is used to recover the camera pose via a PnP solver within a RANSAC loop. The yellow circles in the IFR module denote specific operations. See Sec.~\ref{sec:ifr} and Fig.~\ref{fig:IFR} for details.
    } 
    \label{fig:overview}
    \vspace{-5mm}
\end{figure*}

\subsubsection{\textbf{Data Preprocessing}} Given an initial camera pose $\boldsymbol{T}_{\text{init}}$, we transform the raw LiDAR points $P_w$ into the camera coordinate system and project them onto image plane using a pinhole camera model to obtain the depth map $I_{D}$. Meanwhile, a structurally clear event image within a given time window is generated. The event image generation in~\cite{chen2025evloc} involves two steps: deblurring and denoising. In our implementation, we retain the deblurring step but omit denoising due to its high computational cost. Instead, we apply a Spatio-Temporal Contrast Filter and a Trail Filter from the Prophesee SDK~\cite{prophesee_sdk} to preprocess the raw event stream. These filters effectively reduce both noise and event count. After filtering, we apply the event image generation algorithm from \cite{chen2025evloc} to generate the final event image $I_{EV}$.

\subsubsection{\textbf{Flow Estimation Network}}
Same as~\cite{chen2025evloc}, we formulate the cross-modal event–depth matching problem as a dense event–depth flow prediction task using a modified RAFT~\cite{teed2020raft} network, trained with a masked L2 loss:
\begin{equation}
\mathcal{L}_{\text{flow}}
= \frac{\sum_{u,v} \mathbf{1}\big[\boldsymbol{f_{\text{gt}}}(u,v) \neq 0\big] 
\left\|\boldsymbol{f}(u,v)-\boldsymbol{f_{\text{gt}}}(u,v)\right\|_2}
{\sum_{u,v} \mathbf{1}\big[\boldsymbol{f_{\text{gt}}}(u,v) \neq 0\big] + \epsilon},
\end{equation}
where $\boldsymbol{f}$ is the predicted event-depth flow. $\boldsymbol{f_{\text{gt}}}$ denotes the ground-truth event–depth flow, obtained by projecting 3D points into the image plane using the ground truth pose $\boldsymbol{T}_{\text{gt}}$ and the initial pose $\boldsymbol{T}_{\text{init}}$, and then computing the difference between the corresponding 2D projections.
$\mathbf{1}\big[\boldsymbol{f_{\text{gt}}}(u,v) \neq 0\big]$ returns 1 if the ground-truth flow is valid (non-zero), and 0 otherwise. $\epsilon$ is a small constant to ensure numerical stability.

\textbf{Remarks}: EVLoc~\cite{chen2025evloc} demonstrates the feasibility of using a flow estimation network to predict dense correspondences between events and LiDAR maps. However, the method provides limited analysis of cross-modal consistency between events and LiDAR point clouds. The primary challenge in this task stems from the modality and appearance gap between depth maps and event images, making cross-modal matching substantially more difficult than intra-modal alignment. This naturally suggests that reducing this gap by making depth maps more ``event-like'' could simplify the matching process. Motivated by this insight, we redesign the framework to explicitly enhance cross-modal consistency and thereby further improve localization accuracy. The details of our method are presented in the following section.

\section{Methodology}
In this section, we introduce our novel framework integrating the flow estimator alongside the edge detector (Sec.~\ref{sec:edge}) in a mutually reinforcing manner, which is achieved via two novel cross-task feature fusion modules (Sec.~\ref{sec:cff} and \ref{sec:ifr}). An overview of our framework is illustrated in Fig.~\ref{fig:overview}.

\subsection{Edge Detection Network}
\label{sec:edge}
It's known that events are typically triggered in gradient-rich regions—namely, along edges in event images. This insight motivates the idea of extracting edges from depth maps to facilitate cross-modal alignment. However, depth maps encode only geometric edge structures and lack the texture edge information present in event images, making such alignment not always reliable, particularly in complex or noisy environments. To address this, we propose to learn an encoder-decoder edge detector that extracts edge features from depth, and implicitly enhances feature-level consistency between modalities, as detailed in the Sec.~\ref{sec:cff} and \ref{sec:ifr}.

Due to the lack of learning-based edge detectors specifically designed for depth maps, we adopt the widely-used HED network\cite{xie2015holistically}, originally developed for RGB images. HED is a convolutional network that extracts five feature maps at resolutions $(1, 1/2, 1/4, 1/8, 1/16)$ from intermediate layers. Each feature map is passed through a $1 \times 1$ convolution and upsampled to match the input resolution, producing five side outputs. These outputs are then fused via a $1 \times 1$ convolution to generate the final edge map. 

We use the input event image to generate ground truth edge map. Specifically, since the depth map generated from ground truth pose is aligned with the event image, we first apply dilation to the depth map to ensure coverage of sparse event pixels. Then, for each event pixel, we retrieve its corresponding depth value $d$. Using the pinhole camera model, we reconstruct the corresponding 3D points $P_{edge}$ in the camera coordinate system.
Next, we reproject the 3D edge points using the projection function 
$\pi$, based on the relative transformation between the ground truth pose and the initial pose: 
\begin{equation}
I_{\text{EV}}^{'} \triangleq \pi (P_{\text{edge}}, \boldsymbol{T}_{\text{init}}^{-1}\boldsymbol{T}_{\text{gt}})    
\end{equation}
Finally, we generate the ground truth edge map $I_{\text{ED}}$ by applying binarization to the reprojected event image $I_{\text{EV}}^{'}$. 

The network is trained using a class-balanced cross-entropy loss function \cite{xie2015holistically}, defined as:
\begin{equation}
\begin{aligned}
\mathcal{L}_{\text{edge}} = &\, - \sum_{u,v} \Big[ w  I_{\text{ED}}(u, v) \log \left( p(u, v) \right) \\
&+ (1 - w) (1 - I_{\text{ED}}(u, v)) \log \left( 1 - p(u, v) \right) 
\Big]
\end{aligned}
\end{equation}
Here, $p(u,v)$ denotes the predicted edge probability at pixel $(u,v)$, and $w$ the class-balancing weight calculated based on the proportion of non-edge pixels in the ground truth edge map $I_{\text{ED}}$. This loss addresses the imbalance between edge and non-edge pixels by assigning a higher weight to the minority class (edge pixels).

\begin{figure}[ht]
    \centering
    \includegraphics[width=1.0\linewidth]{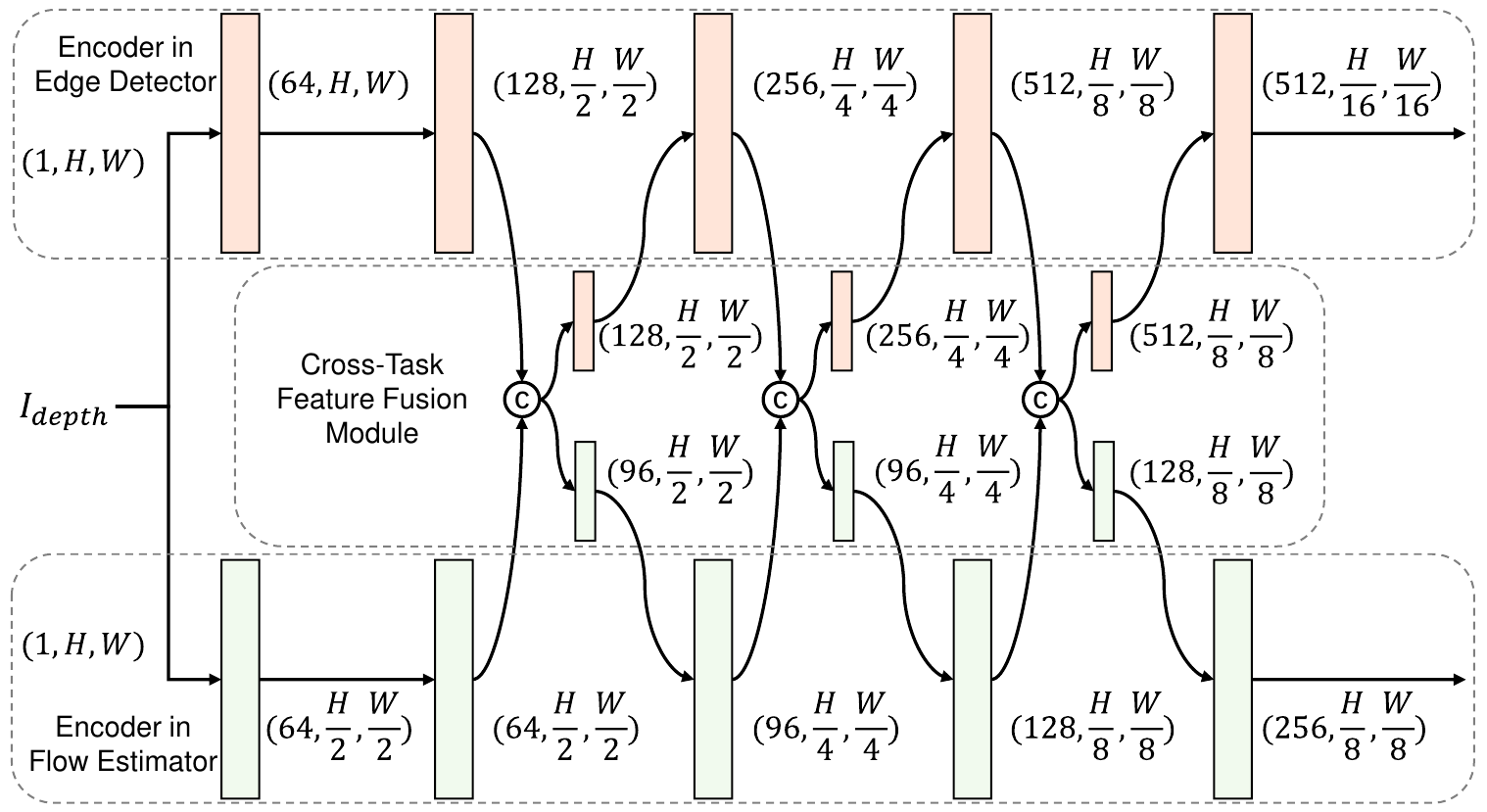}
    \caption{\textit{Architecture of the CFF module.} Both the edge detection and flow estimation branches employ five-layer encoders with different spatial resolutions at each layer. Feature fusion is performed at the middle three layers, as illustrated in the diagram.} 
    \label{fig:Fusion}
    \vspace{-5mm}
\end{figure}

\subsection{Cross-task Feature Fusion}
\label{sec:cff}
\begin{figure}[b]
	\centering
	\begin{minipage}[b]{0.49\linewidth}
		\centering
		\includegraphics[width = 1.0\columnwidth]{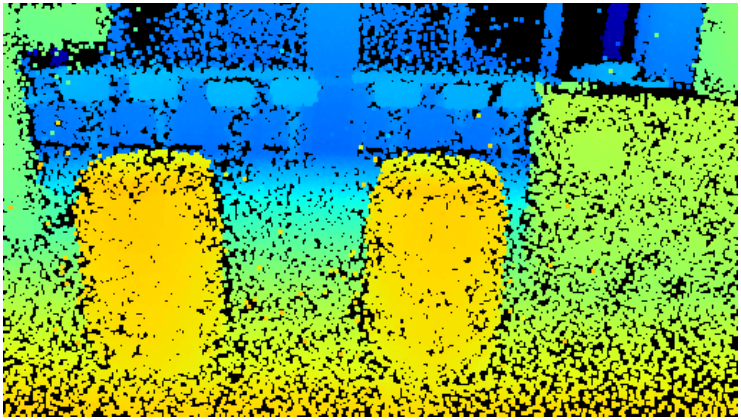} 
		\vspace{1mm}
        \centerline{\small(a) 
        \textit{Depth map}}
	\end{minipage}
	\begin{minipage}[b]{0.49\linewidth}
		\centering
		\includegraphics[width = 1.0\columnwidth]{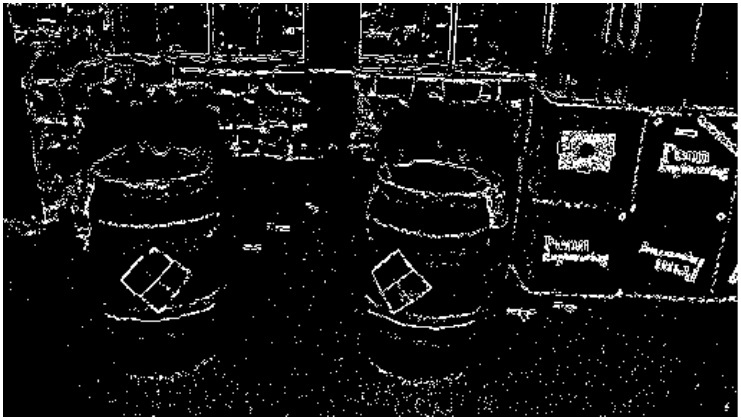} 
		\vspace{1mm}
        \centerline{\small(b) \textit{Ground truth edge map}}
	\end{minipage}
	\begin{minipage}[c]{0.49\linewidth}
		\centering
		\includegraphics[width = 1.0\columnwidth]{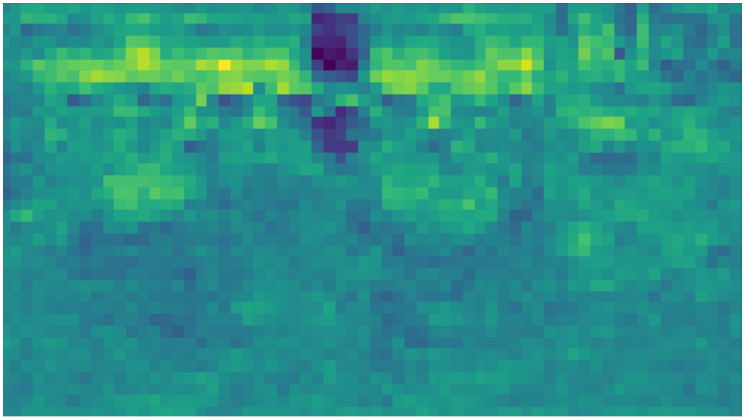} 
		\centerline{\small(c) \textit{Feature map from baseline}}
	\end{minipage}
	\begin{minipage}[d]{0.49\linewidth}
		\centering
		\includegraphics[width = 1.0\columnwidth]{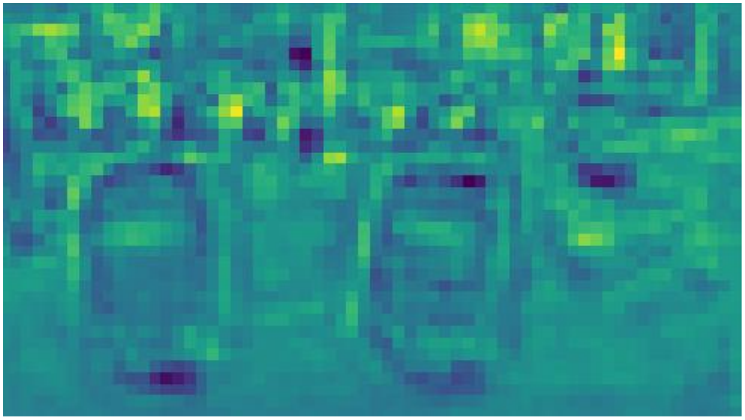} 
		\centerline{\small(d) \textit{Feature map after adding CFF}}
	\end{minipage}
	\caption{\textit{Visualization of the encoded depth feature map.} Each pixel value in the visualized feature map indicates the mean value of the corresponding feature vector at that location. The depth feature map after applying the CFF module contains richer texture details that align with structures.} 
    \label{fig:Feature}
    \vspace{-5mm}
\end{figure}

As outlined above, the generated depth map and event image are provided to the flow estimator and edge detector, which initially produce the event features $F_{EV}$, depth features $F_D$, and edge features $F_{ED}$. The depth encoders in both branches are integrated through our proposed CFF module, as illustrated in Fig.~\ref{fig:Fusion}. Multi-scale feature maps from the encoders of the edge detector and flow estimator are progressively fused and propagated. The flow estimator encoder extracts five feature maps at resolutions $(1/2, 1/2, 1/4, 1/8, 1/8)$ from the input depth map, while the edge detector encoder produces maps at $(1, 1/2, 1/4, 1/8, 1/16)$. Due to the resolution mismatch, fusion is performed only at the middle three layers. Features at layer $i$ are concatenated and passed through two separate $1 \times 1$ convolutions to restore the original channel dimensions for each task branch, formulated as follows: 
\begin{equation}
    \hat{F}_s^i = \text{Conv}_{1 \times 1}^s \big(\text{Concat}(F_{\text{ED}}^i, \, F_{\text{D}}^i)\big),
\end{equation}
where $s \in \{\text{ED}, \text{D}\}$ indicates the task branch (edge detection or depth), $\hat{F}_{\text{ED}}^i$ and $\hat{F}_{\text{D}}^i$ are enhanced features.
Although both encoders operate on the same input depth map, they focus on different aspects: the edge detection encoder emphasizes fine-grained, low-level edge cues, while the flow estimation encoder primarily captures motion-related features, focusing on both local correspondences and global motion context. Through cross-task feature fusion, the depth features gain enhanced sensitivity to local texture details, while the edge features become more geometrically structured, enhancing cross-modal consistency. As illustrated in Fig.~\ref{fig:Feature}, the depth feature map obtained after applying the CFF module contains richer texture details that align with structures present in the ground truth edge map.


\begin{figure}[ht]
    \centering
    \includegraphics[width=0.9\linewidth]{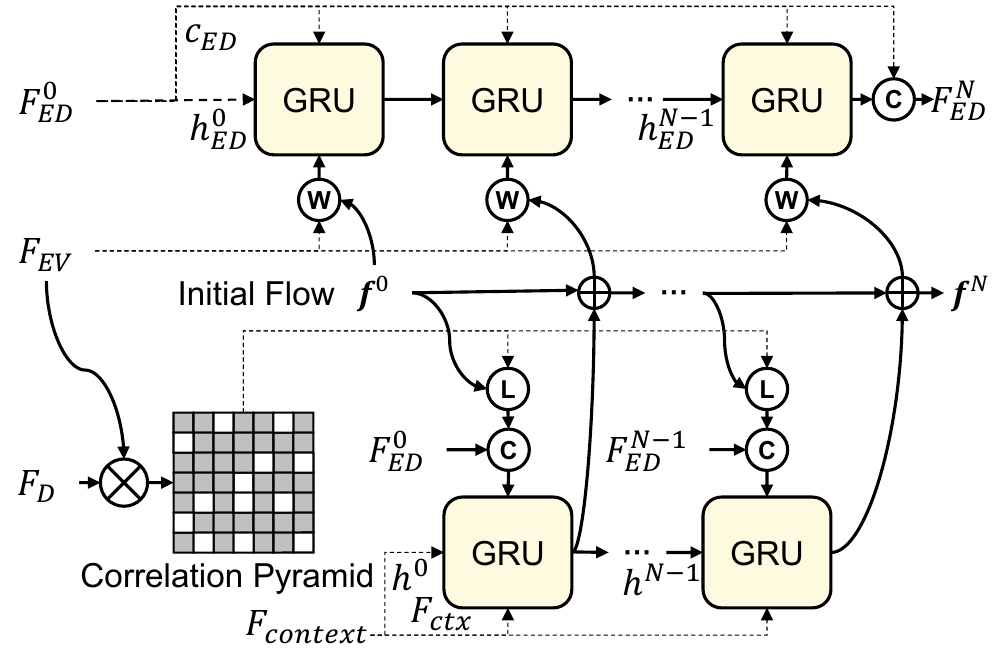}
    \caption{\textit{Architecture of the IFR module.} 
    %
    %
    Edge features ($F_{ED}^{0} \to F_{ED}^{N}$) and flow estimates ($\boldsymbol{f}^{0} \to \boldsymbol{f}^{N}$) are progressively refined through mutual reinforcement. W, L, and C denote warping, indexing and concatenating, respectively.
    } 
    \label{fig:IFR}
    \vspace{-1mm}
\end{figure}

\subsection{Iterative Feature Refinement}\label{sec:ifr}
The obtained features are fed into the IFR module, as shown in Fig.~\ref{fig:IFR}. This module consists of two components: (1) an edge update branch (\textbf{top}) that refines edge features by integrating information from aligned event features, and (2) a flow update branch (\textbf{bottom}) that refines flow estimates by incorporating edge features into the correlation volumes.

We first construct a correlation pyramid by computing the dot product between all pairs of depth features $F_D$ and event features $F_{EV}$. The initial flow $\boldsymbol{f}^0$ is set to zero. 
At iteration $n (n\in[1, N])$, the current flow estimate $\boldsymbol{f}^{n-1}$ is used to index into the correlation pyramid and retrieve a corresponding correlation volume $F_C^n$. In the flow update branch, the correlation volume is first concatenated with the current edge feature $F_{ED}^{n}$ along the channel dimension, then passed through a $1 \times 1$ convolution to restore the original channel size. A GRU cell updates the hidden state using $\hat{F}_C^n$, the current flow, and context features $F_{\text{ctx}}$ (with $F_{\text{ctx}}$ and $h^{0}$ from the EVLoc-style context encoder \cite{chen2025evloc}), and the resulting state is mapped by a flow head $\textit{H}$ to predict an incremental flow update. Then, in the edge update branch, the updated flow is used to warp the event feature $F_{EV}$, spatially aligning it with the edge feature $F_{ED}$. The current edge feature $F_{ED}^n$, extracted from the fourth encoder layer to match the resolution of the event feature, is split into a context feature $c_{ED}$ and a hidden state $h^n_{ED}$. A GRU cell takes as input the context feature, the hidden state, and the aligned event features, and outputs an updated hidden state. The updated hidden state is concatenated with the original context feature to form the refined edge feature $F_{ED}^{n+1}$.
\begin{equation}
\hat{F}_C^n = \text{Conv}_{1\times 1}\big(\text{Concat}(F_C^n, F_{ED}^n)\big),
\end{equation}
\begin{equation}
\boldsymbol{f}^n = \boldsymbol{f}^{n-1} + \textit{H}(\text{GRU}\big(h^{n-1}, \hat{F}_C^n, \boldsymbol{f}^{n-1}, F_{\text{ctx}}\big)),
\end{equation}
\begin{equation}
\resizebox{0.9\linewidth}{!}{$
F_{ED}^{n+1} = \text{Concat}(c_{ED}, \text{GRU}\big(h_{ED}^n, \textit{W}(F_{EV}, \boldsymbol{f}^n), c_{ED}\big)).
$}
\end{equation}

By integrating edge features in this way, the correlation volume is enriched with local structural details, complementing the global motion context and enabling the model to predict finer-grained flow. Notably, the two branches are mutually reinforcing: improved flow leads to better-aligned event features, while more accurate edge features enhance the fused correlation representation, further guiding the flow refinement process. See Fig.~\ref{fig:Edge} the bottom rows. 

\subsection{Other Details}
\textbf{Loss function: } As described above, the edge detector and flow estimator are trained jointly in a mutually reinforcing manner. The final loss is defined as a weighted combination of the two terms:
\begin{equation}
\mathcal{L}=\alpha \times \mathcal{L}_{flow} + \beta \times \mathcal{L}_{edge},
\end{equation}
where $\alpha$ is $1$ and $\beta$ is $100$ in our experiments.

\textbf{Pose estimation: }
We treat pixels with non-zero depth values as valid and extract flow predictions (i.e., 2D-3D correspondences) only at these locations. We use the publicly available PoseLib library~\cite{PoseLib}, with a maximal reprojection error of 12.0 and Huber loss for optimization, while keeping all other parameters at their default values.

\begin{table}[t]
    \centering
    \caption{Ablation study results. Bold numbers indicate the best performance except the upper bound \textit{Oracle}.
    }
    \label{tab:Ablation}
    \resizebox{1.0\linewidth}{!}{
    \begin{tabular}[t]{c|c|c|c|c|c|c|c}
        \toprule
        \toprule
        & \multirow{2}{*}{CFF} & \multirow{2}{*}{IFR} & \multirow{2}{*}{EPE} & \multicolumn{2}{c|}{Mean Error} & \multicolumn{2}{c}{Median Error}\\
        & & & & Transl.[cm] $\downarrow$ & Rot.[°] $\downarrow$ & Transl.[cm] $\downarrow$& Rot.[°] $\downarrow$\\
        \midrule
        Oracle & & & 1.66 & 4.14 & 0.39 & 3.31 & 0.22\\
        \midrule
        Baseline\cite{chen2025evloc} & & & 6.10 & 9.01 & 1.10 & 8.11 & 0.97\\
        \midrule
        Direct Mask & & & 6.11 & 8.99 & 1.02 & 7.80 & 0.85 \\
        \midrule
        (a) & $\checkmark$ & & 5.75 & \textbf{8.06} & 0.94 & 7.20 & 0.83\\
        \midrule
        (b) & $\checkmark$ & $\checkmark$ & \textbf{5.52} & 8.07 & \textbf{0.93} & \textbf{6.85} & \textbf{0.74}\\
        \bottomrule
        \bottomrule
    \end{tabular}
    }
    \vspace{-5mm}
\end{table}
\section{Experiments}

\subsection{Datasets and Implementation Details}
\subsubsection{Datasets} 
M3ED dataset \cite{Chaney_2023_CVPR} is a typical dataset that provides high-definition LiDAR maps and ground truth trajectories, so we primarily evaluate our method on it. For additional evaluation, we incorporate the DSEC dataset\cite{gehrig2021dsec}, which introduces more complex driving scenarios not present in M3ED. However, DSEC only offers disparity images. To facilitate evaluation, we reconstruct a 3D point cloud for each sample by back-projecting the disparity-derived depth into 3D space using the pinhole camera model. Additionally, since our method relies on high-quality, pre-built LiDAR maps to produce dense and accurate depth estimates, we exclude sequences heavily affected by dynamic objects, as their motion introduces significant noisy artifacts into the reconstructed maps.

\subsubsection{Data Processing} 
To simulate uncertainty in coarse localization, we generate an initial pose estimate by perturbing the ground truth (identity matrix for DSEC) through random uniform sampling, with translation offsets within $\pm50$ cm and rotation offsets within $\pm5$°.
To ensure consistency and efficiency, all input images are cropped and resized to fixed resolutions specific to each dataset. For the M3ED dataset, images are resized and cropped to $512 \times 288$; for DSEC, images are cropped to $480 \times 360$. 

\begin{figure}[h]
    \centering
    \includegraphics[width=0.95\linewidth]{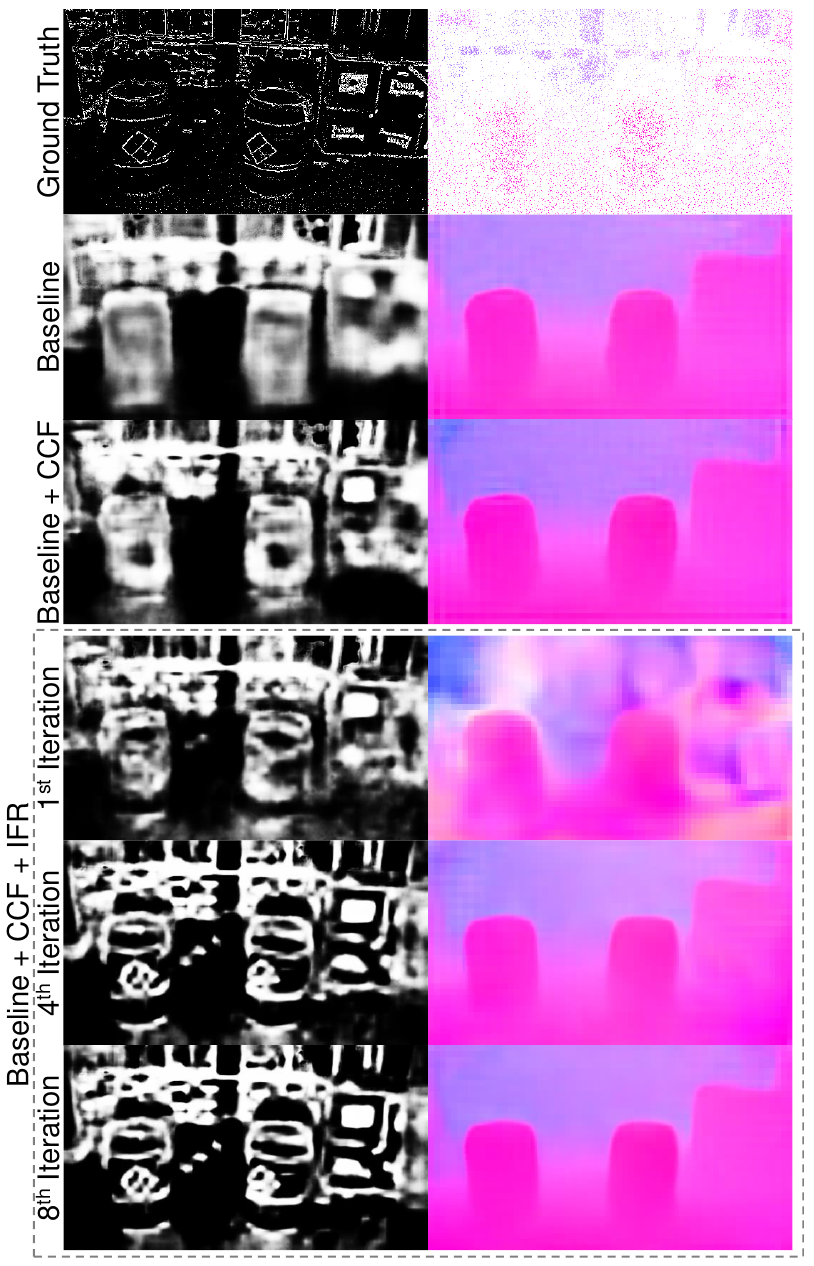}
    \caption{\textit{Visualization of the predicted edges and flows. }\textbf{Left column:} edge predictions; \textbf{Right column:} flow estimates.
    } 
    \label{fig:Edge}
    \vspace{-5mm}
\end{figure}

\subsubsection{Training Details}
For M3ED, a separate model is trained per scene, as the dataset covers heterogeneous environments (indoor flight, outdoor parking, forest, etc.); for example, in the $falcon\_indoor\_flight$ scene, the first two sequences are used for training and the last for evaluation. For scenes with only one sequence (e.g., $spot\_indoor\_building\_loop$), we adopt a temporal split, using the first $70\%$ of data for training and the remaining $30\%$ for evaluation, ensuring that training and testing do not overlap. In contrast, a single model is trained across all DSEC scenes—since they all depict driving environments with similar sensor configurations and motion patterns—with the final sequence of each scene used for evaluation.
In each experiment, the model is trained for 100 epochs with a batch size of 2, weight decay of $1 \times 10^{-4}$, an initial learning rate of $4 \times 10^{-5}$, and a \emph{OneCycleLR} learning rate scheduler, using the Adam optimizer\cite{kingma2014adam}. The IFR module is configured to run for 12 iterations during training and 24 iterations during testing. All experiments are conducted on a single NVIDIA RTX 4090 GPU.
\subsubsection{Evaluation Metrics}
For evaluation, localization accuracy is assessed using the mean and median errors of the predicted poses, while the average End-Point Error (EPE) is used to evaluate the accuracy of the estimated flow.  

\subsection{Ablation Study}
To evaluate the effectiveness of the proposed modules, we conduct a series of ablation experiments, as demonstrated in Table~\ref{tab:Ablation} and Fig.~\ref{fig:Edge}. The \emph{baseline} is EVLoc~\cite{chen2025evloc}, which we retrain using the same image resolution and experimental settings as our method.
In the \emph{oracle} setting, we use the ground truth edge map to mask the depth input before both training and testing, producing a masked depth map that closely resembles the texture of the input event image. This minimizes the appearance gap between modalities. As expected, the model significantly outperforms the baseline, demonstrating the theoretical upper bound of flow-based matching when perfect edge information is available. 
Next, we explore a progressive pipeline \emph{direct mask} where the predicted edge maps are directly used to mask the depth input. The edge detector is trained independently, and an example of the predicted edge map is shown in the 2nd row of Fig.~\ref{fig:Edge}, this configuration slightly outperforms the baseline, validating that even imperfect edge predictions can improve depth–event matching. 

Building upon this, we incrementally incorporate the proposed modules. By fusing edge features into the flow estimator through the CFF module, the \emph{model (a)} achieves further improvements beyond simple masking. This suggests that implicit feature-level integration is more effective than direct masking. Additionally, as shown in the 3rd row of Fig.~\ref{fig:Edge}, the predicted edges exhibit finer details compared to those produced by a standalone edge detector, indicating that the CFF module also enhances the edge detection branch by integrating complementary information. Adding the IFR module yields further gains in localization accuracy (\emph{model (b)}), confirming its contribution to refining flow estimates. 
Compared to a standalone edge detector HED (Fig.~\ref{fig:Edge}, 2nd row), our method produces sharper and more detailed edge predictions, demonstrating the IFR module’s ability to effectively fuse local texture cues from the event features into the edge representations. Furthermore, as iterations progress, both the predicted edges and the refined flow estimates become increasingly complete and accurate, as illustrated in Fig.~\ref{fig:Edge} the last three rows, highlighting the mutually reinforcing nature of the iterative refinement process.


\begin{table*}[htbp]
    \centering
    \caption{Comparison of localization performance among I2D-Loc, EVLoc, and the proposed method on M3ED and DSEC. Median errors in translation and rotation are used as evaluation metrics. Bold numbers indicate the best performance.}
    \label{tab:Benchmark}
    \resizebox{1.0\linewidth}{!}{
    \begin{tabular}[htbp]{ccccccccc}
        \toprule
        \toprule
         \multirow{2}{*}{Dataset} & \multirow{2}{*}{Test Scene} & \multirow{2}{*}{Time} & \multicolumn{2}{c}{\underline{\quad \qquad I2D-Loc\cite{chen2022i2d} \qquad\quad}} & \multicolumn{2}{c}{\underline{\quad \qquad EVLoc\cite{chen2025evloc} \qquad\quad}} & \multicolumn{2}{c}{\underline{\quad \qquad Ours \qquad\quad}} \\
         & & & Transl.[cm]$\downarrow$ & Rot.[°]$\downarrow$ & Transl.[cm]$\downarrow$ & Rot.[°]$\downarrow$ & Transl.[cm]$\downarrow$ & Rot.[°]$\downarrow$ \\
        \midrule
        \multirow{10}{*}{M3ED\cite{Chaney_2023_CVPR}} & \textit{falcon\_indoor\_flight} & - & - & - & 8.11 & 0.97 & \textbf{6.85} & \textbf{0.74}\\
        & \textit{spot\_indoor\_building\_loop} & - & - & - & 12.88 & 2.07 & \textbf{10.94} & \textbf{1.63} \\
        & \textit{falcon\_outdoor\_day\_penno\_parking} & day & - & - & 25.11 & 1.32 & \textbf{22.40} & \textbf{1.18}\\
        & \textit{falcon\_forest\_into\_forest} & day & - & - & 20.46 & 1.87 & \textbf{19.35} & \textbf{1.84} \\
        & \textit{spot\_outdoor\_day\_penno\_short\_loop} & day & - & - & 17.89 & 0.88 & \textbf{14.37} & \textbf{0.74}\\
        & \textit{spot\_forest\_road} & day & - & - & 13.55 & 0.60 & \textbf{12.96} & \textbf{0.57}\\
        & \textit{car\_urban\_day\_penno} & day & - & - & 14.15 & 0.60 & \textbf{11.88} & \textbf{0.56}\\
        & \textit{car\_forest\_into\_ponds} & day & - & - & 19.76 & 0.86 & \textbf{16.23} & \textbf{0.75}\\
        & \textit{spot\_outdoor\_night\_penno\_short\_loop} & night & - & - & 34.85 & 1.71 & \textbf{24.97} & \textbf{1.04}\\
        & \textit{car\_urban\_night\_penno} & night & - & - & 20.84 & \textbf{0.72} & \textbf{19.15} & 0.87\\
        \midrule
        \multirow{14}{*}{DSEC\cite{gehrig2021dsec}} & \textit{interlaken\_00} & dawn/dusk & 10.45 & 0.42 & 7.11 & 0.30 & \textbf{5.56} & \textbf{0.25}\\
        & \textit{thun\_00}  & dawn/dusk & 11.15 & 0.55 & 8.26 & 0.42 & \textbf{8.08} & \textbf{0.41} \\
        & \textit{zurich\_city\_00} & dawn/dusk & 11.30 & 0.52 & 7.94 & 0.37 & \textbf{6.78} & \textbf{0.34} \\
        & \textit{zurich\_city\_01} & dawn/dusk & 9.88 & 0.36 & 7.63 & 0.30 & \textbf{6.15} & \textbf{0.28} \\
        & \textit{zurich\_city\_02} & dawn/dusk & 8.68 & 0.33 & 7.93 & 0.31 & \textbf{6.68} & \textbf{0.27} \\
        & \textit{zurich\_city\_03} & night & 14.01 & 0.74 & 8.79 & 0.61 & \textbf{8.14} & \textbf{0.51}\\
        & \textit{zurich\_city\_04} & dawn/dusk & 11.82 & 0.51 & 7.49 & 0.30 & \textbf{6.16} & \textbf{0.26}\\
        & \textit{zurich\_city\_05} & day & 9.27 & 0.37 & 6.42 & 0.28 & \textbf{5.36} & \textbf{0.24} \\
        & \textit{zurich\_city\_06} & day & 14.26 & 0.62 & 8.12 & 0.39 & \textbf{6.98} & \textbf{0.37}\\
        & \textit{zurich\_city\_07} & day & 11.33 & 0.43 & 7.27 & 0.34 & \textbf{6.25} & \textbf{0.32} \\
        & \textit{zurich\_city\_08} & dawn/dusk & 8.72 & 0.36 & 6.62 & 0.27 & \textbf{5.81} & \textbf{0.26}\\
        & \textit{zurich\_city\_09} & night & 8.05 & 0.32 & 6.38 & 0.29 & \textbf{5.92} & \textbf{0.28}\\
        & \textit{zurich\_city\_10} & night & 11.35 & 0.49 & 8.37 & 0.35 & \textbf{7.35} & \textbf{0.34} \\
        & \textit{zurich\_city\_11} & dawn/dusk & 9.56 & 0.41 & 6.14 & 0.26 & \textbf{5.21} & \textbf{0.23}\\
        \bottomrule
        \bottomrule
    \end{tabular}
    }
    \vspace{-5mm}
\end{table*}

\subsection{Results Analysis}
In this section, we compare our method against the state-of-the-art event–LiDAR localization approach EVLoc~\cite{chen2025evloc} under the same experimental setup, as well as the RGB-based image-to-LiDAR method I2D-Loc~\cite{chen2022i2d}. Results are shown in Table \ref{tab:Benchmark}. M3ED does not provide ground truth poses for standard cameras. Although approximate poses can be obtained via interpolation, the resulting depth maps exhibit significant misalignment with the camera frames. 
In contrast, DSEC provides ground truth disparity maps for each RGB image, allowing us to apply the same ground truth generation method described above for the experiments on standard cameras. 

As shown in Table \ref{tab:Benchmark}, our proposed method consistently outperforms EVLoc across all sequences under different lighting conditions, demonstrating the overall superiority of the proposed approach. On M3ED, in indoor environments ($falcon\_indoor\_flight$ and $spot\_indoor\_building\_loop$), LEAR reduces translation error by roughly $15–20\%$ and rotation error by about $20-25\%$ compared to EVLoc. In more challenging outdoor scenarios, both methods perform worse overall compared to indoors, suggesting that limited or unstructured scene texture makes cross-modal matching more difficult. Nevertheless, our method does not degrade relative to the baseline and still achieves modest gains, indicating that the model continues to learn useful features from complex environments. Nighttime sequences are particularly challenging: even after denoising and deblurring, the generated event frames contain little useful information of the captured scene but substantial noise, and the high resolution of the event camera further amplifies the noise effects. 
Nevertheless, LEAR achieves performance comparable to EVLoc, despite the lack of useful structural information. 

\begin{figure}[htbp]
    \centering
    \includegraphics[width=0.9\linewidth]{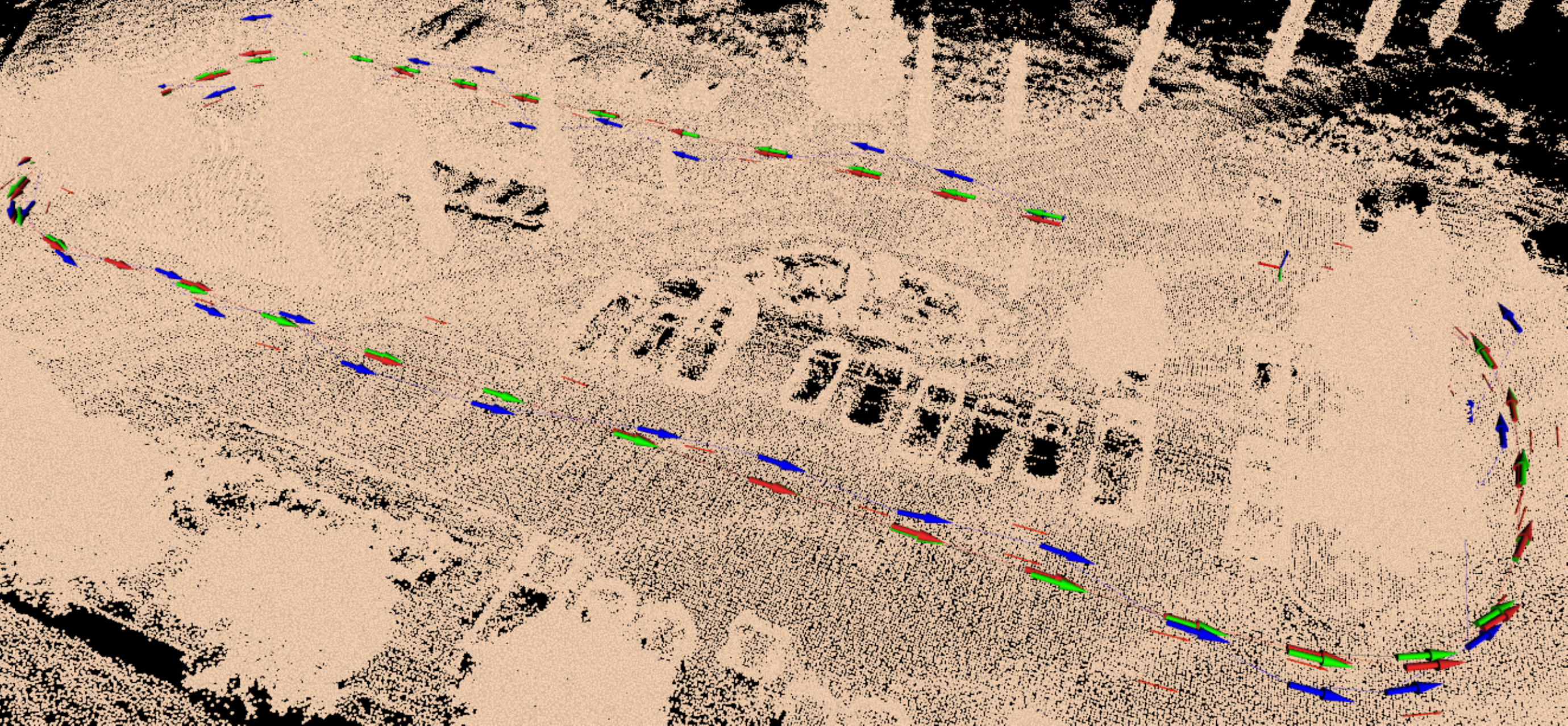}
    \caption{\textit{Comparison of ground-truth poses (green), initial poses (blue), and poses predicted by our method (red) on M3ED.} Our predictions align closely with the ground truth, demonstrating high accuracy.
    } 
    \label{fig:Vis}
    \vspace{-5mm}
\end{figure}

DSEC contains only driving scenarios but under diverse lighting conditions. LEAR consistently outperforms EVLoc and achieves up to $36\%$ lower translation and $29\%$ lower rotation errors, and also surpassing I2D-Loc with reductions of up to $61\%$ in translation and $44\%$ in rotation errors, respectively. 
The significant improvement over I2D-Loc highlights the benefits of event cameras for cross-modal data association.  We believe this could be attributed to the much larger domain gap between daytime and nighttime RGB images, which causes the models to converge to a local minimum. In contrast, event images remain consistent across both daytime and nighttime conditions.


Figure~\ref{fig:Vis} visualizes the localization results of one scene in M3ED. Starting from a coarse initial pose (blue), LEAR refines it to a precise prediction (red) that closely aligns with the ground truth (green).

\begin{table}[htbp]
    \centering
    \caption{Runtime and memory consumption comparison.}
    \label{tab:Runtime}
    \resizebox{\linewidth}{!}{
    \begin{tabular}{l|c|c|c|c}
        \toprule
        \toprule
        Methods & Stage & Runtime [ms] & Memory [MB] & Transl.[cm]/Rot.[°]\\
        \midrule
        \multirow{2}{*}{EVLoc~\cite{chen2025evloc}} 
        & Feature encoding & 4.33 & \multirow{2}{*}{1202} & \multirow{2}{*}{8.11/0.97}\\
        & Flow estimation (24 iters) & 51.80 & &\\
        \midrule
        \multirow{4}{*}{Ours} 
        & Feature encoding & 7.71 & \multirow{2}{*}{1232} & \multirow{2}{*}{6.85/0.74}\\
        & Flow estimation (24 iters) & 100.43 & & \\
        \cline{2-5}
        & Feature encoding & 7.71 & \multirow{2}{*}{1232} & \multirow{2}{*}{7.01/0.75}\\
        & Flow estimation (12 iters) & 51.30 & & \\
        \bottomrule
        \bottomrule
    \end{tabular}
    }
    \vspace{-5mm}
\end{table}

Table~\ref{tab:Runtime} summarizes the runtime and memory consumption of the baseline and our method on M3ED. Under the same number of iterations, our method requires negligible extra cost on memory, while more runtime (primarily in flow estimation) than EVLoc. Nevertheless, this overhead remains practical and enables significant gains in localization accuracy. Furthermore, our method at 12 iterations still outperforms EVLoc at 24 iterations, while they require a comparable runtime in this case. These results demonstrate the effectiveness of the enhanced feature representation and iterative refinement introduced in our framework.
\section{Conclusions and Future Work}
In this paper, we propose a dual-branch framework for event-based visual localization in LiDAR maps. Our approach integrates an optical flow estimation network with an edge detection network, and introduces a cross-task feature fusion module along with an iterative feature refinement module. These components enable implicit feature-level fusion between the two branches, enhancing cross-modal feature consistency in finding accurate correspondences for localization. Extensive experiments on multiple public datasets demonstrate the effectiveness of the proposed method.

\textbf{Limitations:} Despite its advantages, our method relies on local flow estimation in 2D space, which requires sufficient overlap between the input depth map and the event image. Moreover, successful matching depends heavily on the presence of sufficient depth variation. In scenarios with degenerate depth information—such as flat ground or smooth walls—our method may fail due to the lack of distinguishable geometric features. In the future, we plan to investigate direct event-to-point cloud registration without projection, which may offer greater robustness in cases with limited initial pose accuracy or poor geometric variation.

\section{Acknowledgement}
This research project has been supported by the Austrian Science Fund (FWF) under project agreement (I 6747-N) EVELOC.

\bibliographystyle{IEEEtran.bst}
\bibliography{references}

@article{gallego2017event,
  title={Event-based, 6-DOF camera tracking from photometric depth maps},
  author={Gallego, Guillermo and Lund, Jon EA and Mueggler, Elias and Rebecq, Henri and Delbruck, Tobi and Scaramuzza, Davide},
  journal={IEEE Transactions on Pattern Analysis and Machine Intelligence},
  volume={40},
  number={10},
  pages={2402--2412},
  year={2017},
  publisher={IEEE}
}

@inproceedings{pham2020lcd,
  title={Lcd: Learned cross-domain descriptors for 2d-3d matching},
  author={Pham, Quang-Hieu and Uy, Mikaela Angelina and Hua, Binh-Son and Nguyen, Duc Thanh and Roig, Gemma and Yeung, Sai-Kit},
  booktitle={Proceedings of the AAAI conference on artificial intelligence},
  volume={34},
  number={07},
  pages={11856--11864},
  year={2020}
}

@inproceedings{li2021deepi2p,
  title={DeepI2P: Image-to-point cloud registration via deep classification},
  author={Li, Jiaxin and Lee, Gim Hee},
  booktitle={Proceedings of the IEEE/CVF Conference on Computer Vision and Pattern Recognition},
  pages={15960--15969},
  year={2021}
}

@inproceedings{bryner2019event,
  title={Event-based, direct camera tracking from a photometric 3d map using nonlinear optimization},
  author={Bryner, Samuel and Gallego, Guillermo and Rebecq, Henri and Scaramuzza, Davide},
  booktitle={IEEE International Conference on Robotics and Automation},
  pages={325--331},
  year={2019},
  organization={IEEE}
}

@article{zuo2024cross,
  title={Cross-modal Semidense 6-DOF tracking of an event camera in challenging conditions},
  author={Zuo, Yi-Fan and Xu, Wanting and Wang, Xia and Wang, Yifu and Kneip, Laurent},
  journal={IEEE Transactions on Robotics},
  volume={40},
  pages={1600--1616},
  year={2024},
  publisher={IEEE}
}

@inproceedings{yuan2024evit,
  title={EVIT: Event-based visual-inertial tracking in semi-dense maps using windowed nonlinear optimization},
  author={Yuan, Runze and Liu, Tao and Dai, Zijia and Zuo, Yi-Fan and Kneip, Laurent},
  booktitle={IEEE/RSJ International Conference on Intelligent Robots and Systems},
  pages={10656--10663},
  year={2024},
  organization={IEEE}
}

@article{xu2025mets,
  title={METS: Motion-Encoded Time-Surface for Event-Based High-Speed Pose Tracking},
  author={Xu, Ninghui and Wang, Lihui and Yao, Zhiting and Okatani, Takayuki},
  journal={International Journal of Computer Vision},
  pages={1--19},
  year={2025},
  publisher={Springer}
}

@inproceedings{yuan2016fast,
  title={Fast localization and tracking using event sensors},
  author={Yuan, Wenzhen and Ramalingam, Srikumar},
  booktitle={IEEE International Conference on Robotics and Automation},
  pages={4564--4571},
  year={2016},
  organization={IEEE}
}

@article{chen2022i2d,
  title={I2D-Loc: Camera localization via image to lidar depth flow},
  author={Chen, Kuangyi and Yu, Huai and Yang, Wen and Yu, Lei and Scherer, Sebastian and Xia, Gui-Song},
  journal={ISPRS Journal of Photogrammetry and Remote Sensing},
  volume={194},
  pages={209--221},
  year={2022},
  publisher={Elsevier}
}

@InProceedings{Chaney_2023_CVPR,
    author    = {Chaney, Kenneth and Cladera, Fernando and Wang, Ziyun and Bisulco, Anthony and Hsieh, M. Ani and Korpela, Christopher and Kumar, Vijay and Taylor, Camillo J. and Daniilidis, Kostas},
    title     = {M3ED: Multi-Robot, Multi-Sensor, Multi-Environment Event Dataset},
    booktitle = {Proceedings of the IEEE/CVF Conference on Computer Vision and Pattern Recognition Workshops},
    month     = {June},
    year      = {2023},
    pages     = {4015-4022}
}

@article{gehrig2021dsec,
  title={Dsec: A stereo event camera dataset for driving scenarios},
  author={Gehrig, Mathias and Aarents, Willem and Gehrig, Daniel and Scaramuzza, Davide},
  journal={IEEE Robotics and Automation Letters},
  volume={6},
  number={3},
  pages={4947--4954},
  year={2021},
  publisher={IEEE}
}

@inproceedings{lee2021eventvlad,
  title={EventVLAD: Visual place recognition with reconstructed edges from event cameras},
  author={Lee, Alex Junho and Kim, Ayoung},
  booktitle={IEEE/RSJ International Conference on Intelligent Robots and Systems},
  pages={2247--2252},
  year={2021},
  organization={IEEE}
}

@article{fischer2022many,
  title={How many events do you need? event-based visual place recognition using sparse but varying pixels},
  author={Fischer, Tobias and Milford, Michael},
  journal={IEEE Robotics and Automation Letters},
  volume={7},
  number={4},
  pages={12275--12282},
  year={2022},
  publisher={IEEE}
}

@inproceedings{ren2024simple,
  title={A simple and effective point-based network for event camera 6-dofs pose relocalization},
  author={Ren, Hongwei and Zhu, Jiadong and Zhou, Yue and Fu, Haotian and Huang, Yulong and Cheng, Bojun},
  booktitle={Proceedings of the IEEE/CVF Conference on Computer Vision and Pattern Recognition},
  pages={18112--18121},
  year={2024}
}

@misc{PoseLib,
  title = {PoseLib - Minimal Solvers for Camera Pose Estimation},
  author = {Viktor Larsson and contributors},
  URL = {https://github.com/vlarsson/PoseLib},
  year = {2020}
}

@inproceedings{milford2015towards,
  title={Towards visual slam with event-based cameras},
  author={Milford, Michael and Kim, Hanme and Leutenegger, Stefan and Davison, Andrew},
  booktitle={Robotics: Science and System Workshops},
  year={2015}
}

@inproceedings{teed2020raft,
  title={Raft: Recurrent all-pairs field transforms for optical flow},
  author={Teed, Zachary and Deng, Jia},
  booktitle={European Conference on Computer Vision},
  pages={402--419},
  year={2020},
  organization={Springer}
}

@misc{prophesee_sdk,
  author       = {Prophesee},
  title        = {Prophesee Metavision SDK},
  note         = {Available at \url{https://docs.prophesee.ai/stable/index.html}},
}

@article{kingma2014adam,
  title={Adam: A method for stochastic optimization},
  author={Kingma, Diederik P and Ba, Jimmy},
  journal={arXiv preprint arXiv:1412.6980},
  year={2014}
}

@article{jin20216,
  title={A 6-DOFs event-based camera relocalization system by CNN-LSTM and image denoising},
  author={Jin, Yifan and Yu, Lei and Li, Guangqiang and Fei, Shumin},
  journal={Expert Systems with Applications},
  volume={170},
  pages={114535},
  year={2021},
  publisher={Elsevier}
}

@inproceedings{lin20226,
  title={6-dof pose relocalization for event cameras with entropy frame and attention networks},
  author={Lin, Hu and Li, Meng and Xia, Qianchen and Fei, Yifeng and Yin, Baocai and Yang, Xin},
  booktitle={Proceedings of the 18th ACM SIGGRAPH International Conference on Virtual-Reality Continuum and Its Applications in Industry},
  pages={1--8},
  year={2022}
}

@inproceedings{nguyen2019real,
  title={Real-time 6dof pose relocalization for event cameras with stacked spatial lstm networks},
  author={Nguyen, Anh and Do, Thanh-Toan and Caldwell, Darwin G and Tsagarakis, Nikos G},
  booktitle={Proceedings of the IEEE/CVF Conference on Computer Vision and Pattern Recognition Workshops},
  pages={0--0},
  year={2019}
}

@article{kong2022event,
  title={Event-VPR: End-to-end weakly supervised deep network architecture for visual place recognition using event-based vision sensor},
  author={Kong, Delei and Fang, Zheng and Hou, Kuanxu and Li, Haojia and Jiang, Junjie and Coleman, Sonya and Kerr, Dermot},
  journal={IEEE Transactions on Instrumentation and Measurement},
  volume={71},
  pages={1--18},
  year={2022},
  publisher={IEEE}
}

@article{fischer2020event,
  title={Event-based visual place recognition with ensembles of temporal windows},
  author={Fischer, Tobias and Milford, Michael},
  journal={IEEE Robotics and Automation Letters},
  volume={5},
  number={4},
  pages={6924--6931},
  year={2020},
  publisher={IEEE}
}

@INPROCEEDINGS{9635907,
  author={Lee, Alex Junho and Kim, Ayoung},
  booktitle={IEEE/RSJ International Conference on Intelligent Robots and Systems}, 
  title={EventVLAD: Visual Place Recognition with Reconstructed Edges from Event Cameras}, 
  year={2021},
  volume={},
  number={},
  pages={2247-2252},
  doi={10.1109/IROS51168.2021.9635907}}

@article{xing2023target,
  title={Target-free extrinsic calibration of event-lidar dyad using edge correspondences},
  author={Xing, Wanli and Lin, Shijie and Yang, Lei and Pan, Jia},
  journal={IEEE Robotics and Automation Letters},
  volume={8},
  number={7},
  pages={4020--4027},
  year={2023},
  publisher={IEEE}
}

@inproceedings{ta2023l2e,
  title={L2E: Lasers to events for 6-DoF extrinsic calibration of lidars and event cameras},
  author={Ta, Kevin and Bruggemann, David and Br{\"o}dermann, Tim and Sakaridis, Christos and Van Gool, Luc},
  booktitle={IEEE International Conference on Robotics and Automation},
  pages={11425--11431},
  year={2023},
  organization={IEEE}
}

@inproceedings{xie2015holistically,
  title={Holistically-nested edge detection},
  author={Xie, Saining and Tu, Zhuowen},
  booktitle={Proceedings of the IEEE/CVF International Conference on Computer Vision},
  pages={1395--1403},
  year={2015}
}

@article{hussaini2024applications,
  title={Applications of spiking neural networks in visual place recognition},
  author={Hussaini, Somayeh and Milford, Michael and Fischer, Tobias},
  journal={IEEE Transactions on Robotics},
  year={2024},
  publisher={IEEE}
}

@article{claxton2024improving,
  title={Improving Visual Place Recognition Based Robot Navigation By Verifying Localization Estimates},
  author={Claxton, Owen and Malone, Connor and Carson, Helen and Ford, Jason J and Bolton, Gabe and Shames, Iman and Milford, Michael},
  journal={IEEE Robotics and Automation Letters},
  year={2024},
  publisher={IEEE}
}

@article{chen2025evloc,
  title={EVLoc: Event-based Visual Localization in LiDAR Maps via Event-Depth Registration},
  author={Chen, Kuangyi and Zhang, Jun and Fraundorfer, Friedrich},
  journal={arXiv preprint arXiv:2503.00167},
  year={2025}
}

@article{liu2024gs,
  title={GS-EVT: Cross-Modal Event Camera Tracking based on Gaussian Splatting},
  author={Liu, Tao and Yuan, Runze and Ju, Yi'ang and Xu, Xun and Yang, Jiaqi and Meng, Xiangting and Lagorce, Xavier and Kneip, Laurent},
  journal={arXiv preprint arXiv:2409.19228},
  year={2024}
}

@inproceedings{nair2024enhancing,
  title={Enhancing Visual Place Recognition via Fast and Slow Adaptive Biasing in Event Cameras},
  author={Nair, Gokul B and Milford, Michael and Fischer, Tobias},
  booktitle={IEEE/RSJ International Conference on Intelligent Robots and Systems},
  pages={3356--3363},
  year={2024},
  organization={IEEE}
}

@article{lin2023e2pnet,
  title={E2pnet: event to point cloud registration with spatio-temporal representation learning},
  author={Lin, Xiuhong and Qiu, Changjie and Shen, Siqi and Zang, Yu and Liu, Weiquan and Bian, Xuesheng and M{\"u}ller, Matthias and Wang, Cheng and others},
  journal={Advances in Neural Information Processing Systems},
  volume={36},
  pages={18076--18089},
  year={2023}
}
\end{document}